\documentclass[10pt,twocolumn,letterpaper]{article}

\usepackage{wacv}              % To produce the CAMERA-READY version
\usepackage{graphicx}
\usepackage{siunitx}
\usepackage{booktabs}
\usepackage{amsmath}
\usepackage{amssymb}
\usepackage{booktabs}
\usepackage[pagebackref,breaklinks,colorlinks]{hyperref}

\usepackage[capitalize]{cleveref}
\crefname{section}{Sec.}{Secs.}
\Crefname{section}{Section}{Sections}
\Crefname{table}{Table}{Tables}
\crefname{table}{Tab.}{Tabs.}

\def\wacvPaperID{3}
\def\confName{WACV}
\def\confYear{2025}

\begin{document}

\title{Snapshot: Towards Application-centered Models for Pedestrian Trajectory Prediction in Urban Traffic Environments}

\author{Nico Uhlemann, Yipeng Zhou, Tobias Simeon Mohr, Markus Lienkamp\\
Technical University of Munich, Germany; School of Engineering \& Design, Institute\\
of Automotive Technology and Munich Institute of Robotics and Machine Intelligence\\
{\tt\small nico.uhlemann@tum.de}
}
\maketitle

%%%%%%%%% ABSTRACT
\begin{abstract}
This paper explores pedestrian trajectory prediction in urban traffic while focusing on both model accuracy and real-world applicability. While promising approaches exist, they often revolve around pedestrian datasets excluding traffic-related information, or resemble architectures that are either not real-time capable or robust. To address these limitations, we first introduce a dedicated benchmark based on Argoverse 2, specifically targeting pedestrians in traffic environments. Following this, we present Snapshot, a modular, feed-forward neural network that outperforms the current state of the art, reducing the Average Displacement Error (ADE) by \SI{8.8}{\percent} while utilizing significantly less information. Despite its agent-centric encoding scheme, Snapshot demonstrates scalability, real-time performance, and robustness to varying motion histories. Moreover, by integrating Snapshot into a modular autonomous driving software stack, we showcase its real-world applicability. \footnote{\url{https://github.com/TUMFTM/Snapshot}}
\end{abstract}

%%%%%%%%% BODY TEXT
\section{Introduction}
\label{sec:intro}

The reliable prediction of pedestrian behavior plays a vital role across various disciplines. Most prominently, it has been extensively studied to understand crowd behavior \cite{koh_modeling_2011, dutra_gradient-based_2017, ondrej_synthetic-vision_2010, van_den_berg_interactive_2008, van_den_berg_reciprocal_2011, kothari_human_2022}, to replicate realistic trajectories \cite{gupta_social_2018, yuan_agentformer_2021, amirian_social_2019, scholler_flomo_2021}, and to enable autonomous systems to consider human motion \cite{salzmann_trajectron_2021, ridel_literature_2018, achaji_is_2022, cadena_pedestrian_2022, fridovich-keil_confidence-aware_2020}. Autonomous driving combines all of these research efforts with the goal of enhancing the safety of the vulnerable road users involved. As this technology integrates more and more into the intricate urban landscape where humans and vehicles closely coexist, reliably predicting pedestrian behavior has proven to be a significant challenge. Despite considerable advancements in recent years, many methods still revolve around well-established trajectory prediction benchmarks such as ETH/UCY \cite{lerner_crowds_2007, pellegrini_youll_2009} and SDD \cite{robicquet_learning_2016}, which primarily contain pedestrians in non-traffic settings. Although specialized datasets such as Argoverse 2 \cite{wilson_argoverse_2023} or nuScenes \cite{caesar_nuscenes_2020} address this shortcoming, they are rarely utilized in pedestrian research due to their primary focus on vehicles or all road users combined. Moreover, the developed algorithms typically prioritize benchmark performance over minimizing runtime or enhancing model robustness, which hinders their applicability in real-world settings. To this end, the insights from pedestrian-centered research, such as the importance of certain features \cite{dutra_gradient-based_2017, uhlemann_evaluating_2024} or approaches themselves \cite{yue_human_2022, zhang_forceformer_2023, mangalam_goals_2020}, remain unused.

To address these aspects, the presented work investigates pedestrian prediction in urban traffic scenarios while combining the knowledge collected from different disciplines. We identified the most crucial features that influence prediction performance and show that both applicability and state-of-the-art prediction accuracy can be achieved with a single model. The contributions of this work can be summarized as follows:
\begin{itemize}
    \item \textbf{Approach:} We introduce Snapshot, a modular, non-recursive approach to real-world pedestrian trajectory prediction that leverages existing work on transformer architectures \cite{kosaraju_social-bigat_2019, gilles_home_2021, cheng_forecast_2023} in combination with Convolutional Neural Networks (CNNs) \cite{mangalam_goals_2020, zamboni_pedestrian_2022}. %Our method achieves state-of-the-art performance while considering as little as two observed timesteps and spatial information only.
    \item \textbf{Feature Analysis:} Based on our model, we evaluate different modifications alongside an ablation study to show the effectiveness of our architecture and determine the significance of individual inputs.
    \item \textbf{Benchmark:} To support future research in the field, we provide a dedicated pedestrian-focused benchmark derived from Argoverse 2 \cite{wilson_argoverse_2023}. It offers a specialized development platform with more than one million training, validation, and test samples combined. % and is meant to track both progress and insights.
    \item \textbf{Applicability:} Lastly, we verify Snapshot's applicability by integrating it into an autonomous driving software stack to gather insights about its real-world performance.
\end{itemize}

\section{Related Work}
\label{sec:rel_work}

\subsection{Pedestrian Datasets}
\label{subsec:datasets}
\begin{figure*}[tb]
  \centering
  \begin{subfigure}{0.49\linewidth}
    \centering
    \includegraphics[height=5.0cm]{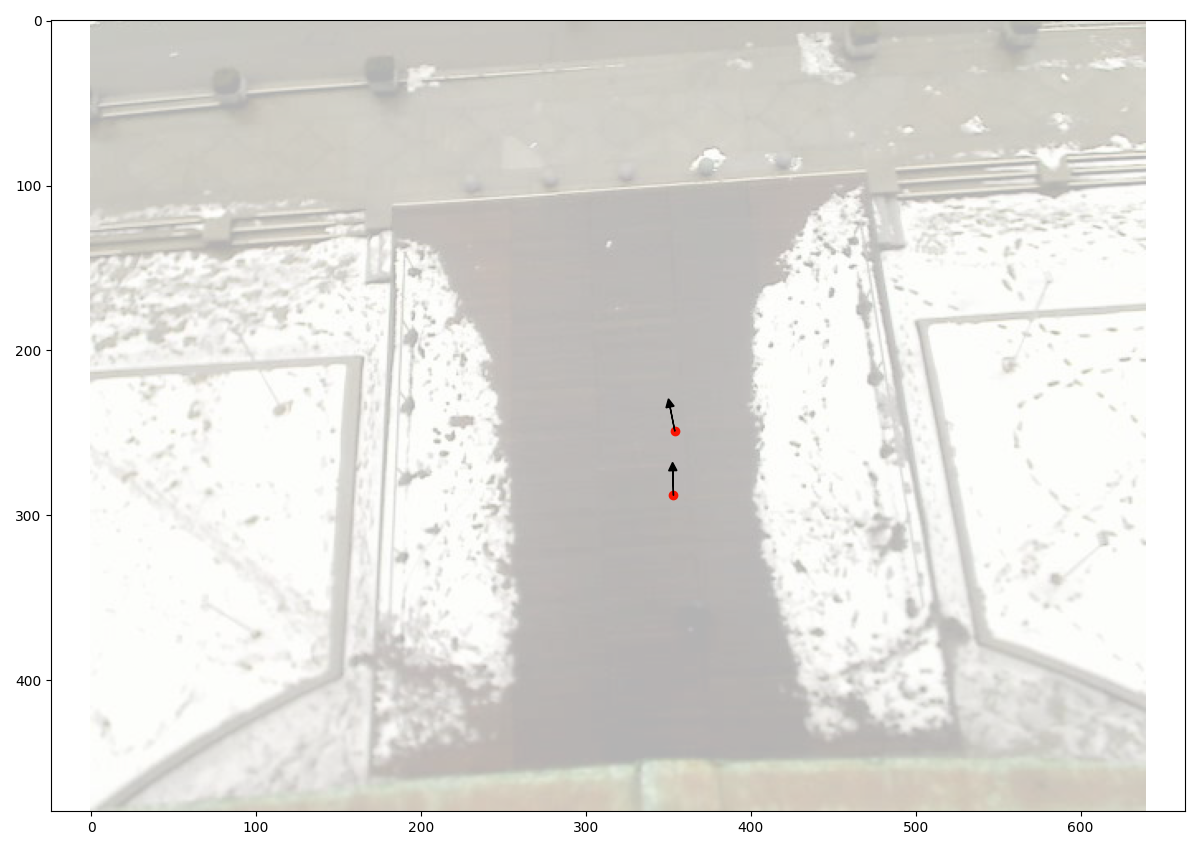}
    \caption{Scene from the ETH dataset \cite{lerner_crowds_2007} focusing only on pedestrians. The scene context is visualized by the faint background image, showing the entrance area of a building during winter.}
    \label{fig:eth_scene}
  \end{subfigure}
  \hfill
  \begin{subfigure}{0.49\linewidth}
    \centering
    \includegraphics[height=5.0cm]{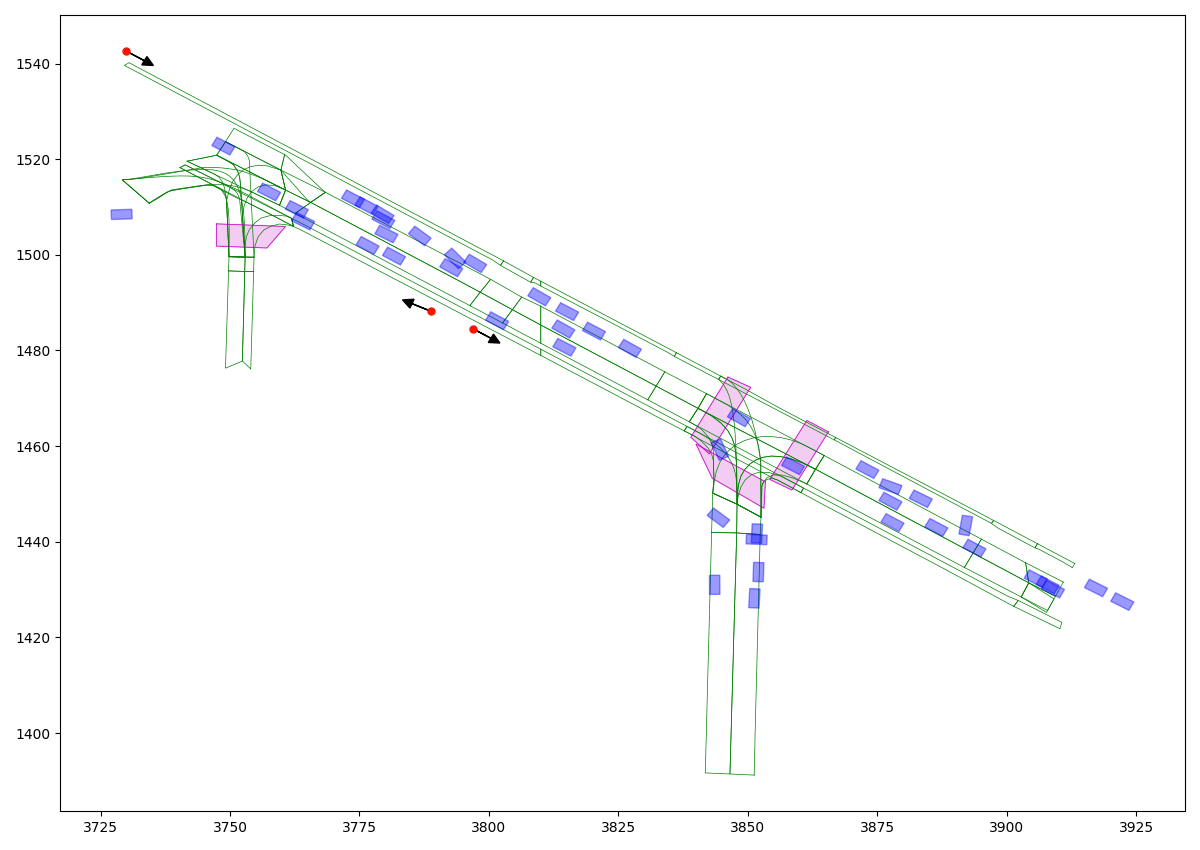}
    \caption{Scene from the Argoverse 2 dataset \cite{wilson_argoverse_2023}. The road network is represented by lanelets with green lines and crosswalks in purple. Vehicles are shown by blue rectangles.}
    \label{fig:av2_scene}
  \end{subfigure}
  \caption{Comparison showcasing a pedestrian-only scenario on the left in contrast to a traffic environment scene on the right. In both cases, pedestrians are highlighted with red circles and back heading arrows.}
  \label{fig:pedestrian_vs_traffic}
\end{figure*}

%1. Abschnitt: Datasets
Comprehensive datasets are the foundation for every learning-based approach. Within the field of pedestrian trajectory prediction, these can be divided into two categories depicted in \cref{fig:pedestrian_vs_traffic}: Pedestrian-only and urban traffic environments. Pedestrian-only datasets comprise outdoor as well as indoor settings, with the popular ETH/UCY \cite{lerner_crowds_2007, pellegrini_youll_2009} and SDD \cite{robicquet_learning_2016} dataset being recorded in urban environments, whereas others like ATC \cite{brscic_person_2013} and Thör \cite{rudenko_thor_2020} were captured within buildings. While this category focuses on establishing a better understanding of how pedestrians move and interact with one another and their environment, the category of urban traffic environments contains vehicles alongside pedestrians and is therefore better suited to explore the overall dynamics concerning autonomous systems. To date, the most popular datasets in this category are comprised of nuScenes \cite{caesar_nuscenes_2020}, Argoverse 2 \cite{wilson_argoverse_2023} and the Waymo Open Motion dataset \cite{ettinger_large_2021}. All of these have in common that the prediction task is conducted based on a birds-eye view (BEV) of the scene, where the recorded tracks, as well as semantic information, are provided in a 2D plane. Since these datasets are based on real-world recordings from a vehicle perspective, common phenomena such as occlusions, tracking losses, and detection inaccuracies are contained within the data. Therefore, they provide an advantage when training networks for real-world applications \cite{sun_human_nodate, hagenus_survey_2024}. 

\subsection{Prediction Approaches}
Historically, pedestrian behavior has been imitated through knowledge-based methods like the Social Force Model \cite{helbing_social_1995} or velocity-based approaches \cite{kim_brvo_2015}. Since then, research has shifted to data-driven models where behavior is learned based on prerecorded data \cite{korbmacher_review_2021}. While recurrent neural networks (RNNs) have initially been adopted to capture the interactions unfolding over the observed motion history of each agent \cite{alahi_social_2016, gupta_social_2018, sadeghian_sophie_2018, kosaraju_social-bigat_2019, salzmann_trajectron_2021}, competitive performance has been achieved in recent years by feed-forward architectures \cite{mangalam_goals_2020, mohamed_social-implicit_2022, zamboni_pedestrian_2022}. Most recently, with the success in natural language processing \cite{vaswani_attention_nodate} as well as computer vision \cite{dosovitskiy_image_2021}, transformer models also have been applied to the field of pedestrian trajectory prediction with great success \cite{yuan_agentformer_2021, zhang_forceformer_2023, salzmann_robots_2023}. Similarly, when predicting other road users besides pedestrians, transformer architectures have since been established as state-of-the-art models which underlines their capability to effectively process heterogeneous inputs \cite{zhou_query-centric_2023, nayakanti_wayformer_2023, zhang_simpl_2024, cheng_forecast_2023}. Despite the introduction of various methods to either improve model performance or effectiveness, such as specialized attention mechanisms \cite{yuan_agentformer_2021, zhang_real-time_2023}, efficient input representations \cite{lan_sept_2023, wang_prophnet_2023}, parallel predictions \cite{zhou_hivt_2022, mohamed_social-implicit_2022}, or hybrid approaches \cite{zhang_forceformer_2023, yue_human_2022}, no model exists that can be easily deployed to an autonomous system. As mentioned previously, reasons for that can be found in extensive preprocessing efforts, high inference times, or robustness against varying motion histories. Moreover, it is often unknown how well these models can anticipate pedestrian motion. To explore this aspect, we will evaluate our proposed method alongside the current state of the art.

\subsection{Incorporated features}
\label{subsec:features}

\begin{table}[b]
  \caption{Common observation and prediction horizons for pedestrian-related datasets in urban environments}
  \label{tab:pred_timesteps}
  \centering
  \begin{tabular}{c|c|c|c}
    \toprule
     Dataset & $T_{o}$ in s & $T_{p}$ in s & $f_{s}$ in Hz \\
     \midrule
     ETH/UCY \cite{lerner_crowds_2007, pellegrini_youll_2009} & 3.2 & 4.8 & 2.5\\
     SDD \cite{robicquet_learning_2016} & 3.2 & 4.8 & 2.5\\
     nuScenes \cite{caesar_nuscenes_2020} & 2 & 6 & 2 \\
     Argoverse 2 \cite{wilson_argoverse_2023} & 5 & 6 & 10 \\
     Waymo Open Motion \cite{ettinger_large_2021} & 1 & 8 & 10 \\
     \bottomrule
  \end{tabular}
\end{table}

Most methods consider three major aspects to accurately predict pedestrian trajectories: motion history, semantic information, and interactions \cite{sadeghian_sophie_2018, salzmann_trajectron_2021}. The motion history usually consists of the observed positional values for each agent that are used to determine an individual's behavior. While this information is generally considered valuable, different opinions exist on how much information is necessary. This aspect is highlighted in \cref{tab:pred_timesteps}, listing the observation as well as the prediction horizons for the different datasets introduced in \cref{subsec:datasets}. Within the table, it can be noted that apart from the observation length $T_{o}$ varying between \numrange{1}{5} \SI{}{\second}, the prediction horizon $T_{p}$ ranges between \numrange{4.8}{8} \SI{}{\second}, as does the sampling rate $f_{s}$ with values between \numrange{2}{10} \SI{}{\hertz}. Recent work suggests that at least for pedestrians, an observation sequence exceeding \SI{1}{\second} might not be as relevant \cite{uhlemann_evaluating_2024}, aligning with the challenge provided by the Waymo Open Motion dataset \cite{ettinger_large_2021}.

The second feature influencing the prediction accuracy is semantic or map information, where rasterized representations have been widely used in the past \cite{mangalam_goals_2020, salzmann_trajectron_2021, sadeghian_sophie_2018, gilles_home_2021}. This changed with the introduction of VectorNet \cite{gao_vectornet_2020}, where polylines are directly encoded as vectors instead of discretizing the BEV scene through a grid. Since then, vector representations have gained increasing interest due to their efficient encoding of the environment, especially in combination with graph neural networks and transformer architectures \cite{nayakanti_wayformer_2023, zhou_hivt_2022, lan_sept_2023}. Both representations initially adopted an agent-centric encoding scheme, leading to good results due to rotational and translational invariance. Regardless, this scheme usually suffers from poor scalability with an increasing number of agents to be predicted. For this reason, scene-centric approaches are being developed, allowing for improved scalability and hardware utilization \cite{zhou_query-centric_2023}. 

Last but not least, interactions among road users play a crucial role in predicting one's action \cite{golchoubian_pedestrian_2023}. Different mechanisms to consider these have been tested in the past, with pooling techniques \cite{gupta_social_2018}, graph representations \cite{salzmann_trajectron_2021, mohamed_social-implicit_2022} and attention \cite{yuan_agentformer_2021, sadeghian_sophie_2018} being the most prominent ones. Drawing inspiration from the field of interactive crowd simulation, other features exist that are believed to more closely match human interaction in the real world \cite{ondrej_synthetic-vision_2010, dutra_gradient-based_2017}. Here, the time-to-closest-approach (ttca) and distance-to-closest-approach (dca), as well as the bearing angle are considered to avoid collisions. These approaches work seemingly well in a simulation environment but haven't been employed in the pedestrian prediction literature to the best of our knowledge. Therefore, we will explore the impact of this interaction scheme while comparing it with a purely distance-based selection, currently being among the most effective approaches \cite{zhang_real-time_2023}.
\section{Methodology}
\label{sec:meth}

In this section, we formulate the problem statement and design constraints influencing the overall setup of our approach. Afterwards, details about the preprocessing are provided and the architecture of Snapshot is introduced. Lastly, we define the metrics used to quantify the accuracy of the models in \cref{sec:res}.

\subsection{Problem formulation}
\label{subsec:prob_form}
The problem of predicting pedestrian trajectories is framed as follows: Given a 2D map of the surrounding scene and a sequence of observed positions $Y_o=[p_0, p_1, ..., p_{T_o}]$ for $N$ agents, predict the most likely trajectory $Y_p=[p_{T_o+1},p_{T_o+2},$ $..., p_{T_p}]$ of the focal agent. Both sequences are represented by Cartesian coordinates $p_t = (x_t, y_t)$, indicating an agent's location within the scene at timestep $t$. For the focal agent, a maximum of $T_o=10$ timesteps or \SI{1}{\second} is considered, while the observations for other agents might only be partially present. In accordance with the Argoverse 2 motion forecasting challenge, the prediction horizon is set to $T_p = 60$ timesteps or \SI{6}{\second}.

We argue that for a practical application only the most likely trajectory is important in the short term, resembling a persons action of crossing the street in front of the ego vehicle or not \cite{kato_autoware_2018}. Although pedestrian behavior is often considered multimodal, research on the ETH/UCY dataset indicates that multimodal predictors do not improve the overall accuracy when only the most likely trajectory is considered \cite{uhlemann_evaluating_2024}. Given this constraint, we'll explore the potential of unimodal predictions in urban traffic environments with respect to the state of the art.

\subsection{Dataset and benchmark}
\label{subsec:dataset_generation}

\begin{figure}[]
    \centering
    \def\svgwidth{\linewidth}
    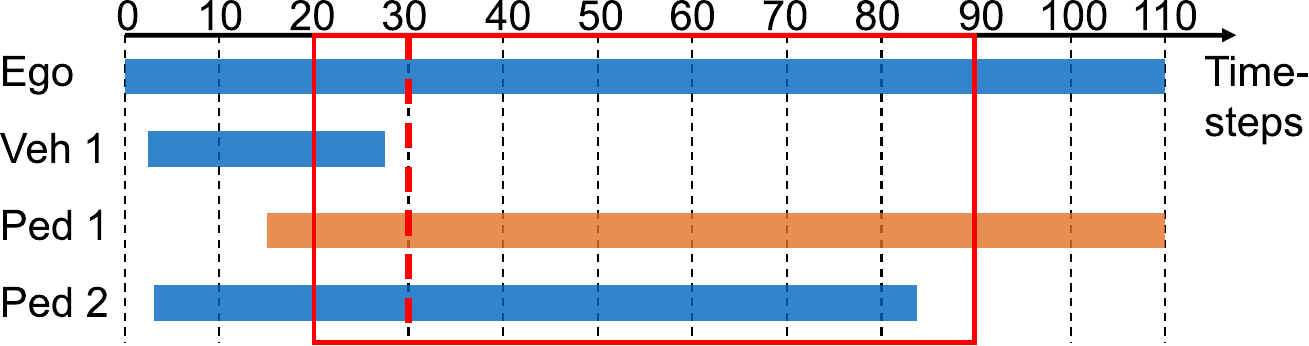
    \caption{Sampling process performed through a sliding window approach, spanning 70 timesteps as visualized by the red bounding box. Pedestrian 1 (Ped 1) is marked as \textit{SCORED\_TRACK} as indicated by the orange color, while the remaining tracks are only employed during observation. Here, Vehicle 1 (Veh 1) is marked as \textit{TRACK\_FRAGMENT} and pedestrian 2 (Ped 2) as \textit{UNSCORED\_TRACK}.}
    \label{fig:sampling}
\end{figure}

For this study, the Argoverse 2 dataset is selected as it contains by far the highest amount of predictable pedestrians when compared to the other options referenced in \cref{subsec:datasets}. Comprising over 250,000 scenes, each lasting \SI{11}{\second}, the dataset encompasses diverse situations and agent types, including vehicles, pedestrians, and cyclists. However, the use of the provided focal tracks for training and evaluation in the single-agent case reveals limitations. Due to the restriction to a \SI{5}{\second} observation length, numerous tracks are excluded from both training and validation. Therefore, only 1,572 pedestrian tracks are available in the validation set. Since prior studies as well as other benchmarks indicate that observations exceeding \SI{1}{\second} might not be relevant for the prediction task \cite{ettinger_large_2021, uhlemann_evaluating_2024}, $T_o$ was limited to that interval. This allows to utilize the provided data more effectively as a sliding window approach can be applied to generate additional samples. This is illustrated in \cref{fig:sampling} with the two displayed pedestrian tracks, which are now processed but would previously been excluded. The sampling process is visualized with the red sliding window, encompassing both observation and ground truth separated by a red, dashed line.

For our approach, pedestrian 1 is marked as predictable, while pedestrian 2 is not due to the track ending within the red sliding window. Despite this, the observable timesteps between 20 and 30 are still included for ego, vehicle 1 and pedestrian 2 as they provide important context to model interactions. This process is performed for each scenario and a sliding interval of five timesteps, dividing it into several predictable scenes. Through this procedure, we generate over one million, \SI{7}{\second} long predictable pedestrian samples for the original training and validation splits combined. Given that the Argoverse 2 challenge evaluates all agent types combined, with pedestrians being only a small fraction of it, we decided to split the generated samples once more to form a dedicated pedestrian benchmark while preserving the original 80-10-10 split ratio. Since over 40\% of the samples contain several pedestrian tracks, the training and evaluation of parallel prediction methods is supported. To better differentiate predictable pedestrian tracks, we adapted the original track labels for these samples. \textit{SCORED\_TRACK} now refers to predictable tracks while \textit{UNSCORED\_TRACK} represents agents still being present in the most recent timestep but either having an incomplete motion history or ground truth. All remaining ones are labeled as \textit{TRACK\_FRAGMENT}. 

\subsection{Feature representation}
\label{subsec:feature}

\begin{figure}[]
    \centering
    \def\svgwidth{\linewidth}
    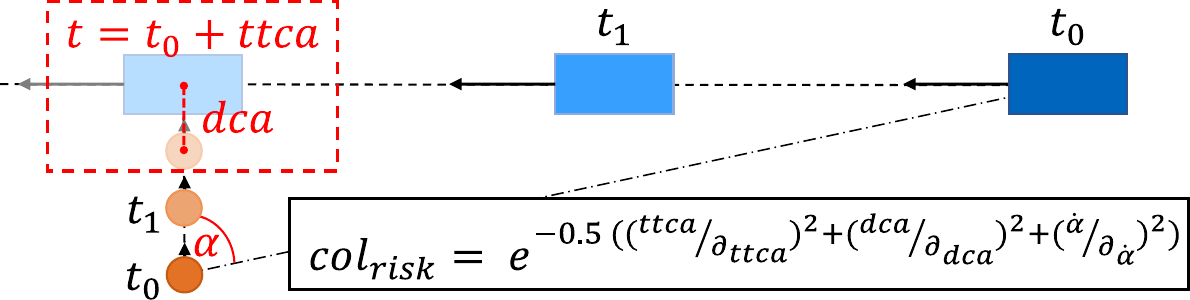
    \caption{Interaction features derived from crowd research. The scene shows a vehicle and the focal pedestrian moving towards one another. Given a constant velocity assumption, the time- and distance-to-closest-approach as well as the derivative of the bearing angle $\dot{\alpha}$ are calculated and used to determine the collision risk $col_{risk}$ as defined by the displayed equation.}
    \label{fig:interaction_features}
\end{figure}

% motion history
Based on the introduced benchmark, features are extracted to improve the model's learning process. Starting with the social features, a matrix with dimensions $8\times21$ is created. In pursuit of incorporating only information relevant to the prediction task, the first dimension of the matrix is defined by Miller’s Law \cite{miller_magical_1956}. It states that individuals can only effectively process around $7\pm2$ objects within a short period of time, limiting the number of agents the focal pedestrian interacts with to seven. Given the possibility that more than seven agents are contained within the scene, relevant ones are determined through a distance-based selection \cite{zhang_real-time_2023}, choosing only the closest agents. To determine the effectiveness of this selection scheme, we compare it to an approach derived from crowd research based on collision risk \cite{dutra_gradient-based_2017}. As can be seen in \cref{fig:interaction_features}, the risk is determined based on the current velocity vectors of the surrounding agents, calculating the time- and distance-to-closest-approach as well as the derivative of the bearing angle $\dot{\alpha}$. The reasoning behind this choice is to include faster-moving objects that are further away, which otherwise might not be considered during a pure distance-based selection. For the calculation, the three parameters are given by $\delta_{ttca} = 1.8$, $\delta_{dca} = 0.3$, and $\delta_{\dot{\alpha}} = 2.0$ and are taken from the initial publication \cite{dutra_gradient-based_2017}.

Besides the number of interacting agents, the second dimension in the social matrix represents the feature vector length shown in \cref{eq:vect_feature}, containing the motion history of an agent for up to ten timesteps $p_t$. As can be seen in the equation, while the first entry indicates the agent type, the remaining twenty only contain relative positional values to simplify preprocessing efforts. In addition, adopting an agent-centric encoding scheme, all agent positions are transformed according to the aligned, local map of the current scene where the origin is defined by the current position $p_{f}$ of the focal pedestrian.

\begin{equation}
    \left [type_{i},\ p_{T_o} ,\ p_{T_o - 1} - p_{f},\ ...\ ,\ p_1 - p_{f},\ p_0 - p_{f}  \right ]
\label{eq:vect_feature}
\end{equation}

\begin{figure}[b]
    \centering
    \def\svgwidth{\linewidth}
    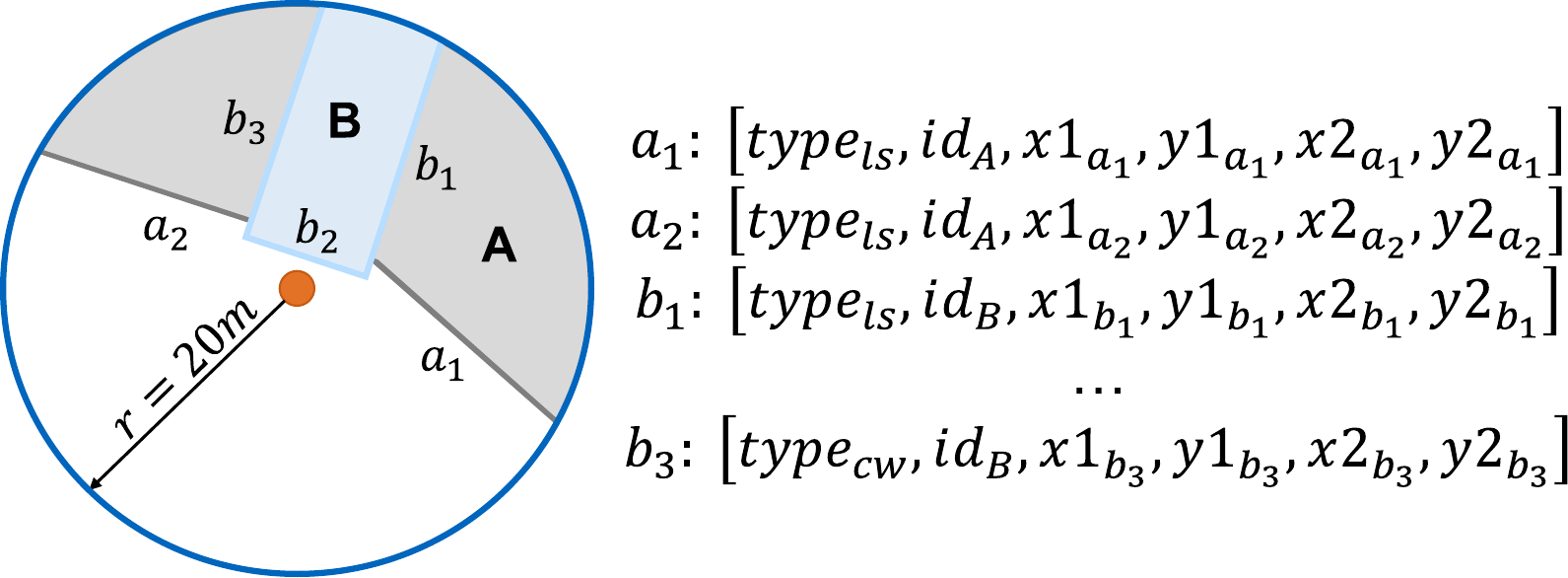
    \caption{Vectorized, local map with a radius $r$ of \SI{20}{\meter} centered around the focal pedestrian. Each polygon, represented by lane segment A and crosswalk B, comprises individual polylines labeled with small letters. For the input, each is transformed into a feature vector as shown on the right, where the first two entries indicate semantic type and polygon id, while the remaining four define the start- and endpoint coordinates.}
    \label{fig:map_encoding}
\end{figure}

%map
While networks employing either agent- or scene-centric encodings can achieve similar results, the former offers better generalization capabilities due to its rotational and translational invariance \cite{zhang_real-time_2023, zhou_hivt_2022}. This effect is particularly noticeable when combined with smaller network sizes. Furthermore, since less semantic information needs to be included for individuals, faster processing times can be achieved in conjunction with a vectorized scene encoding. For our approach a relatively small map feature vector of size $100\times6$ is employed, which is visualized in \cref{fig:map_encoding} alongside our scene representation. For each focal agent, we consider the surrounding polylines within a \SI{20}{\meter} range, measured using the L2 distance. The lines are then ordered, and only the 100 closest ones are integrated into the feature vector. To differentiate various segments, each feature vector consists of six values. The first entry describes the semantic type of the vector, being driveable area, lane segment, crosswalk, or free space, and the second one contains the polygon id this specific line belongs to. The remaining four entries are defined by the individual start and end coordinates, aligned with the agent's heading.

\subsection{Model architecture}
\label{subsec:model}
\begin{figure}[]
    \centering
    \def\svgwidth{0.9\linewidth}
    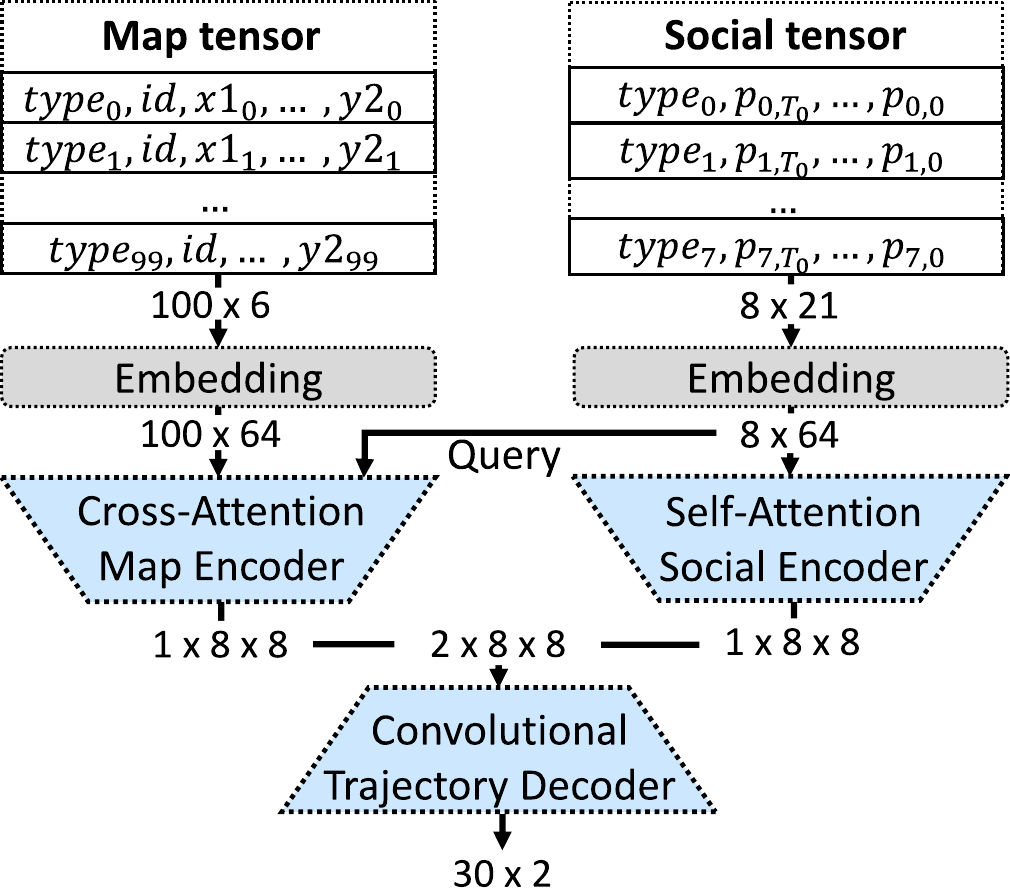
    \caption{Overview of the proposed Snapshot architecture, featuring two independent encoders for social and map information. The subsequent trajectory decoder fuses the information to produce an unimodal output.}
    \label{fig:model_arch}
\end{figure}

In this section, we present our Snapshot architecture. Similar to previous approaches, we opt for a parallel encoding strategy by employing two separate encoders for social and semantic information, as depicted in \cref{fig:model_arch} \cite{salzmann_trajectron_2021, lan_sept_2023, wang_prophnet_2023}. The first encoder processes the social information by extracting relevant features resembling interactions in the scene. Since many publications demonstrate the effectiveness of employing transformer architectures for this type of information \cite{yuan_agentformer_2021, zhou_hivt_2022}, we adopt this approach but modeled it after the original layout \cite{vaswani_attention_nodate}. In our case, the embedding is created through a single fully-connected layer. For the map encoder, we use the same architecture to extract spatial features from the vector map, but exchange the self-attention module with a cross-attention one. Here, the social embedding is provided as query while using the map embedding as key and value \cite{zhou_query-centric_2023}. Both encoding stages produce an output tensor of size $1\times8\times8$ that is concatenated along the channel dimension before being fed into the subsequent decoder. This last stage employs a CNN architecture for mainly two reasons: First, these networks show promising results particularly in pedestrian-centered research \cite{mangalam_goals_2020, zamboni_pedestrian_2022, mohamed_social-implicit_2022}, and second, it allows for the generation of an unimodal trajectory within a single inference pass. To minimize the output dimension, we generate only 30 timesteps with an interval of \SI{0.2}{\second}, interpolating the remaining steps in between. In our experiments, forecasting all 60 timesteps primarily caused the model to learn noise from the ground truth data without significant performance gains. For all three stages, we chose a Leaky ReLU activation function. With this configuration, the presented model has an overall size of just 140,000 parameters.

\subsection{Training procedure}
\label{subsec:training}

As real-world observations are often limited due to an initial detection or temporary occlusions, an accurate predictions needs to be generated for a variable number of observed timesteps to guarantee pedestrian safety. To achieve this robust and accurate performance, the training was conducted in two stages. First, to optimize for accuracy, all available timesteps were considered. In the second phase, individual positional values are removed with a defined ratio for each batch to encourage the model to generalize across various observation lengths, ultimately enhancing robustness. In both cases, we utilize the Average Displacement Error (ADE) as the loss function and upsample the model's output to match the 60 ground truth positions. Additionally, we employ the Adam optimizer with an initial learning rate of \SI{0.0001}, which is automatically adjusted when a plateau during training is reached. To prevent the model from overfitting, L2 regularization with a weight decay of \SI{0.0005} is employed. The training and evaluation of the model were conducted on a single NVIDIA Tesla V100 GPU with 16GB of RAM and a batch size of 256. On average, the training concluded after 60 epochs.

\subsection{Metrics}
\label{subsec:metrics}

The two most common metrics when evaluating trajectory predictors are the Average Displacement Error and the Final Displacement Error (FDE). In the multimodal case where several trajectories are predicted for a given agent, a Best-of-N (BoN) approach is adopted, choosing the trajectory with the smallest error compared to the ground truth sequence $Y_{gt}=[p_{T_o+1}, p_{T_o+2}, ..., p_{T_p}]$. For the scope of this study, only the most likely trajectory is considered to allow for a fair comparison. While the ADE describes the average Euclidean distance between the predicted trajectory $Y_p$ and the ground truth $Y_{gt}$ over the prediction horizon $T_p$, the FDE only considers the last positional value. Hence, the latter can be considered as a measure of the error accumulation over time. Both metrics can be calculated as shown in \cref{eq:ADE} and \cref{eq:FDE}, where $N_p$ represents the number of predicted agents across all scenarios.

\begin{equation}
    ADE = \frac{1}{N_p*T_p} \sum_{i=1}^{N_p} \sum_{t=1}^{T_p} \left|Y_{p}^t \left[ i \right] - Y_{gt}^t \left[ i \right] \right|
\label{eq:ADE}
\end{equation}

\begin{equation}
    FDE = \frac{1}{N_p} \sum_{i=1}^{N_p} \left|Y_{p}^{T_p} \left[ i \right] - Y_{gt}^{T_p} \left[ i \right] \right|
\label{eq:FDE}
\end{equation}

\section{Results}
\label{sec:res}

In the following, we outline the performance of Snapshot by presenting the influence of different training strategies as well as comparing its overall accuracy against the current state of the art. To this end, we will demonstrate the effectiveness of our feature selection. 

\subsection{Training strategy}
\label{subsec:feat_rel}

\begin{figure}[]
    \centering
    \includegraphics[width=\linewidth]{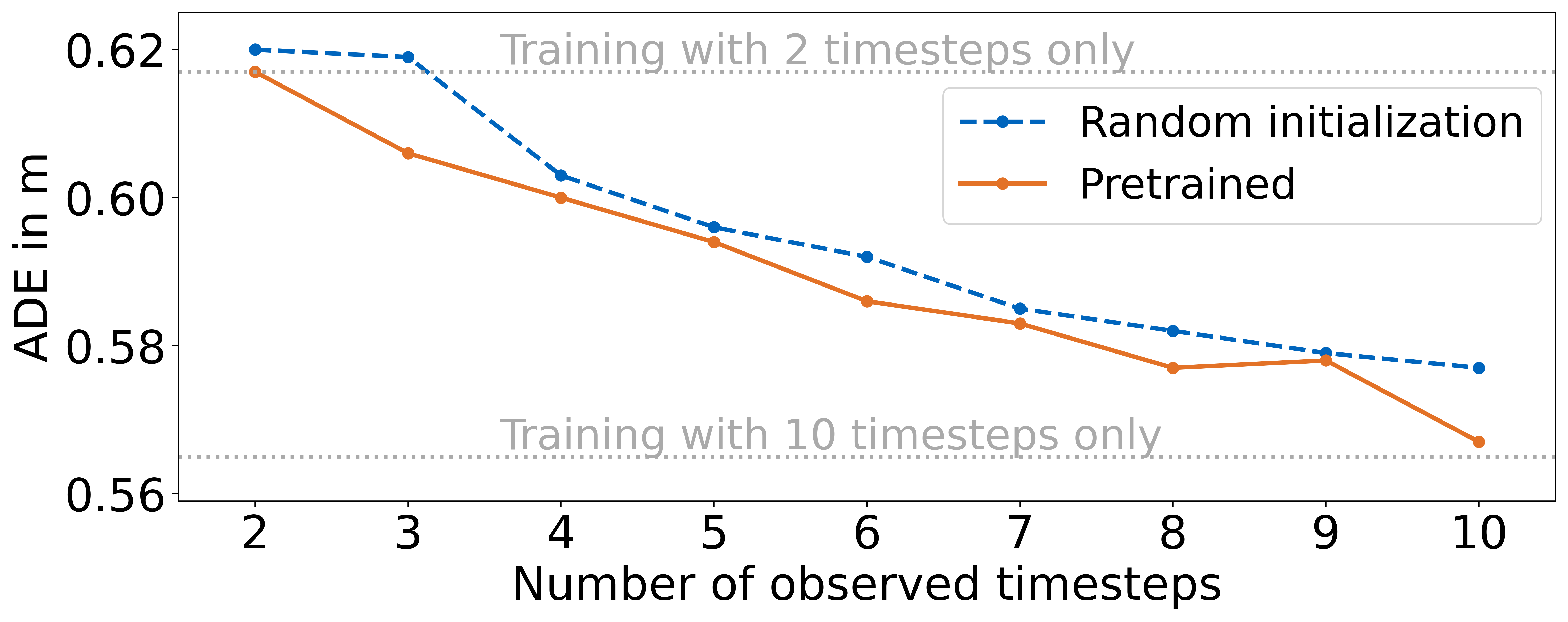}
    \caption{Average Displacement Error for different training strategies and observed timesteps. The blue line resembles training with randomly initialized weights, while the orange one presents our Snapshot training strategy as described in \cref{subsec:training}.}
    \label{fig:res_training}
\end{figure}

An effective training strategy is crucial for achieving robustness alongside accurate predictions as illustrated in \cref{fig:res_training}. Here, we present four distinct methods with varying motion histories. The first two, depicted with dotted horizontal lines, show the accuracy of models optimized exclusively for either two or ten timesteps. These reference values represent the performance obtained when evaluating each model with the corresponding motion history employed during training. When using fewer than ten timesteps for the second, its accuracy significantly deteriorates with ADE and FDE values up to $2.14$ and $3.98$, respectively. To enhance robustness and consider an arbitrary number of timesteps, the colored graphs resemble the remaining two strategies. The blue one illustrates the results when training with random initialization, achieving ADE values between $0.620$ and $0.577$. However, this performance falls short of both baselines for the respective number of timesteps. To leverage previous training runs, the orange curve illustrates the results of our training strategy based on the lower baseline as described in \cref{subsec:training}. This approach shows a similar downward trend compared to random initialization but achieves lower ADE values across all timesteps. Moreover, it demonstrates an almost linear decline that matches both baseline scores. This difference is especially pronounced for ten timesteps, achieving an ADE of $0.567$ and thus maintaining the model's performance from the previous training.

\subsection{Quantitative results}
\label{subsec:quant_res}

To quantify the overall accuracy of Snapshot, we compare its performance against QCNet \cite{zhou_query-centric_2023}, SIMPL \cite{zhang_simpl_2024} and Forecast-MAE \cite{cheng_forecast_2023} which have been selected due to their strong performance in the Argoverse 2 Motion Forecasting competition. As a baseline, we use the constant velocity model (CVM) which has achieved competitive results in previous studies \cite{uhlemann_evaluating_2024}. To conduct a comprehensive evaluation, we present all scores on both, the original Argoverse 2 validation set for focal pedestrians, as well as on the test split for our proposed benchmark.

\begin{table*}[]
  \caption{ADE and FDE values reported on the original Argoverse 2 validation set (1,572 tracks) and our benchmark test split (126,996 tracks), evaluating only the most likely predictions. Lower values indicate better performance.}
  \label{tab:quant_res}
  \centering
  \begin{tabular}{l|cc|cc}
    \toprule
    & \multicolumn{2}{c|}{Argoverse 2 (validation split)} & \multicolumn{2}{c}{Proposed benchmark (test split)} \\
     Models & \hspace{0.3cm}ADE in m\hspace{0.3cm} & \hspace{0.3cm}FDE in m\hspace{0.3cm} & \hspace{0.5cm}ADE in m\hspace{0.5cm} & \hspace{0.5cm}FDE in m\hspace{0.5cm} \\
    \midrule
    CVM \cite{uhlemann_evaluating_2024} & 0.719 & 1.668 & 0.793 & 1.776 \\
    SIMPL \cite{zhang_simpl_2024} & 0.686 & 1.548 & 0.699 & 1.557 \\
    Forecast-MAE \cite{cheng_forecast_2023} & 0.668 & 1.465 & 0.698 & 1.435 \\
    QCNet \cite{zhou_query-centric_2023} & 0.654 & 1.474 & 0.693 & 1.474 \\
    Snapshot(2 timesteps) & 0.648 & 1.423 & 0.617 & 1.342 \\
    Snapshot(10 timesteps) & \textbf{0.605} & \textbf{1.358} & \textbf{0.567} & \textbf{1.251} \\
    \bottomrule
  \end{tabular}
\end{table*}

For the scores of the Argoverse 2 validation split listed in \cref{tab:quant_res} on the left, the baseline CVM reaches an ADE of $0.719$ and an FDE of $1.668$, the overall lowest accuracy among the investigated models. Considering both, 50 historical timesteps as well as semantic information, SIMPL, Forecast-MAE and QCNet manage to significantly improve these values, decreasing the ADE by up to \SI{5.5}{\centi \meter} and the FDE by \SI{19.4}{\centi \meter}. Nevertheless, the overall best results are achieved by Snapshot, achieving ADE and FDE values of up to $0.605$ and $1.358$, respectively. An identical observation can be made for the results of our proposed test split on the right, where the performance differences are even more significant. Here, relative ADE and FDE improvements of \SI{12.6}{\centi \meter} and \SI{18.4}{\centi \meter} can be noted when evaluating Snapshot against the next best model. These results suggest that a \SI{1}{\second} observation window in connection with a unimodal predictor is sufficient to capture the scene dynamics, which will be further discussed in \cref{sec:disc}.

\subsection{Effectiveness of selected features}
\label{subsec:feat_rel}

After having presented the overall performance of our proposed model, this final section briefly compares the influence of various sizes for the local map and different agent selection mechanisms, as described in \cref{subsec:feature}. Starting with the number of polylines considered per agent, the top part of \cref{tab:ablation} displays the ADE and FDE values when considering between \numrange{0}{200} vectors. Generally, a downward trend can be observed, leading to a decrease in accuracy with fewer polylines considered. Completely ablating the map information results in an FDE increase of about \SI{20}{\centi \meter} compared to the overall best result, highlighting the importance of incorporating semantic features. In contrast, using 100 instead of 200 polylines has a negligible effect as a maximum difference of \SI{0.4}{\centi \meter} is observed.

When focusing on social information, being displayed in the bottom row of \cref{tab:ablation}, the distance-based selection mechanism in Snapshot is compared with the collision risk criteria and a complete ablation. It is noticeable that the results improve with the risk-based selection, achieving ADE and FDE values of $0.548$ and $1.196$, respectively. In contrast, when ablating the positional values of the surrounding agents completely, the performance decreases by up to \SI{4}{\centi \meter} in FDE. These aspects highlight that social information contributes valuable cues to generate accurate predictions and that the selection of surrounding agents based on collision risk offers a noticeable advantage.

\begin{table}[]
  \caption{Overview of Snapshot's accuracy when adapting the feature vectors alongside a complete ablation. The displayed tests are conducted on the test set of our proposed benchmark.}
  \label{tab:ablation}
  \centering
  \begin{tabular}{cc|c|c}
    \toprule
    Feature & Variant & ADE in m & FDE in m \\
    \midrule
      & 200 & \textbf{0.563} & \textbf{1.242}\\
     Number of & 100 & 0.565 & 1.246\\ 
     map vectors & 50 & 0.574 & 1.268 \\
      & 0 & 0.647 & 1.444 \\
    \midrule
    Agent se- & L2 Norm & 0.565 & 1.246 \\
    lection & Collision Risk & \textbf{0.548} & \textbf{1.196} \\
    criteria & No agents & 0.583 & 1.286 \\
    \bottomrule
  \end{tabular}
\end{table} 
\section{Discussion}
\label{sec:disc}

In this final section, we will discuss the applicability of Snapshot and provide qualitative results to underline its performance.

\subsection{Applicability}
\label{subsec:appl}

\begin{figure}[]
    \centering
    \includegraphics[width=\linewidth]{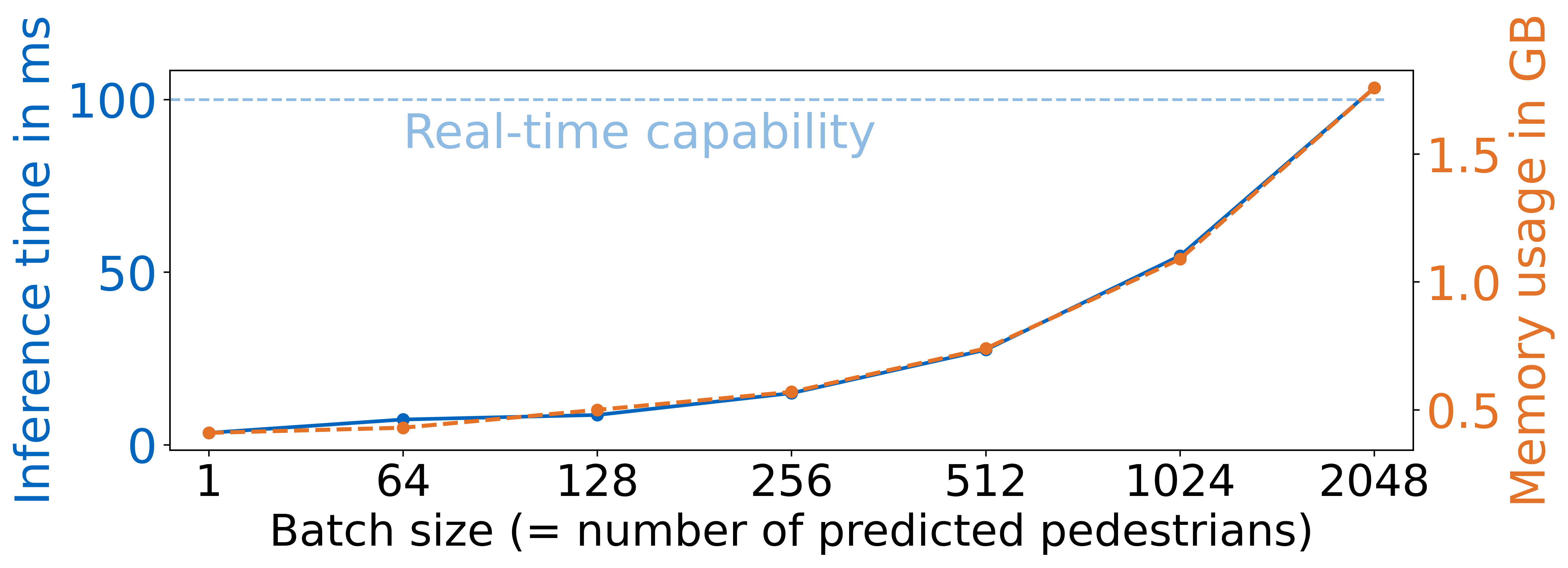}
    \caption{Model inference times in blue alongside the associated GPU memory utilization in orange for different batch sizes. All tests were conducted on a NVIDIA Tesla V100}
    \label{fig:runtime}
\end{figure}

Given the hardware constraints in real-world systems, runtime and computational efficiency are significant challenges when integrating any prediction method. Neglecting the effects of parallel processing first, our model has an average inference time of \SI{3.5}{\milli\second} with a memory utilization of \SI{0.41}{\giga \byte}, marking the lower bound in \cref{fig:runtime}. Due to its relatively small model size, the batch size can be increased to 2,048 before exceeding the real-time mark of \SI{100}{\milli \second}, achieving an overall throughput of \SI{103.44}{\milli\second} at \SI{1.76}{\giga \byte}. Since such agent counts are rarely relevant in the real world and this value is well above the maximum number of 73 pedestrians observed in a single Argoverse 2 scenario, a batch size of 128 represents a suitable choice, resulting in a total prediction time of \SI{8.68}{\milli \second}. Therefore, Snapshot can be considered a real-time capable and scalable architecture.

\begin{figure}[t]
    \centering
    \includegraphics[width=0.95\linewidth]{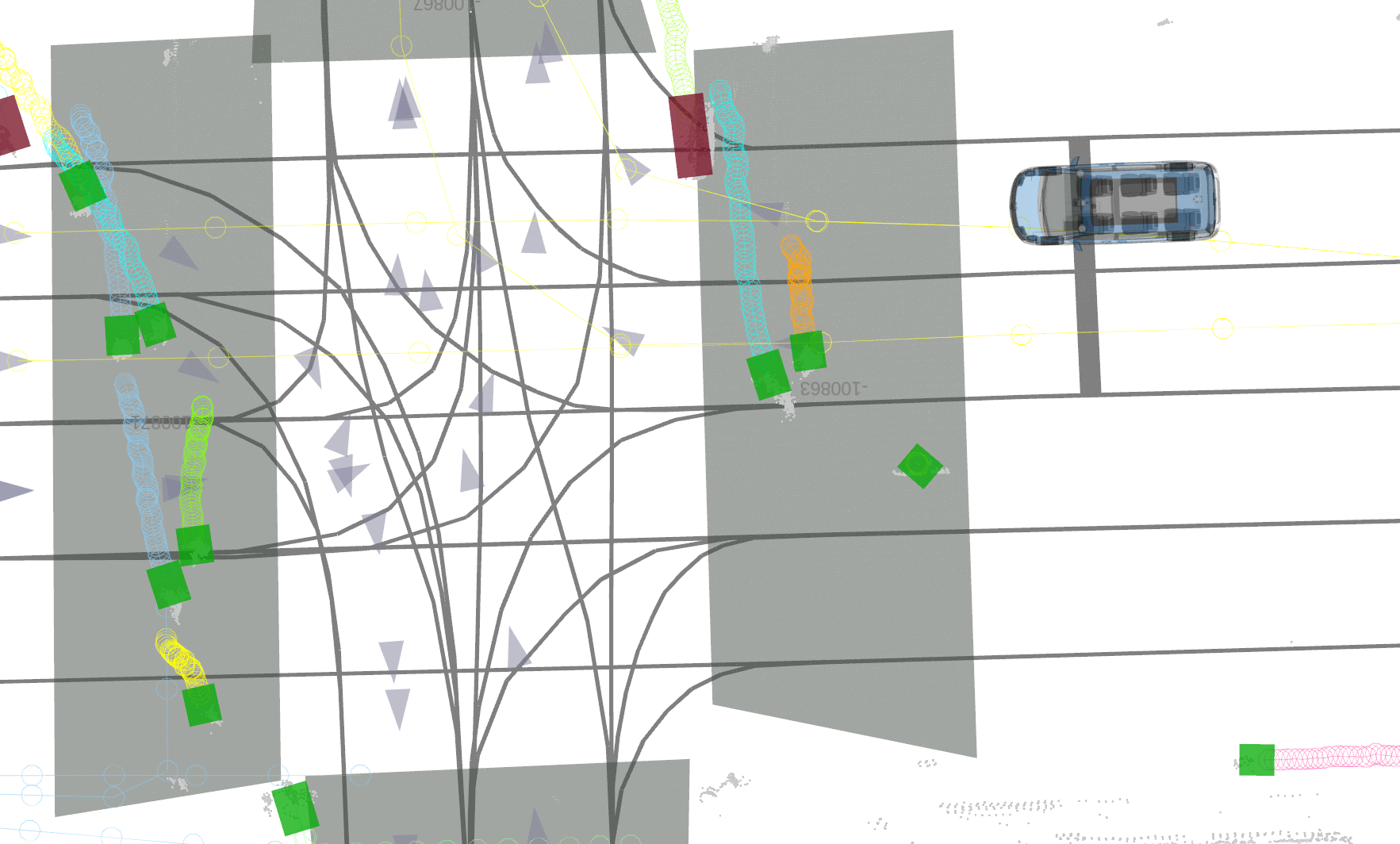}
    \caption{Real-world scenario displaying an intersection with crosswalks. Pedestrians are detected using CenterPoint \cite{yin_center_2021} and highlighted with green rectangles. The predictions are visualized through colored trajectories and are generated by Snapshot running on an automated vehicle as part of a modular software stack.}
    \label{fig:integration}
\end{figure}

To verify that our approach also works in a real-world application and not just in an offline setting, we integrated Snapshot into a modular autonomous driving software stack based on Autoware \cite{kato_autoware_2018} and deployed it onto an automated vehicle. One scene encountered on the road is visualized in \cref{fig:integration} and shows several pedestrians crossing the road over two crosswalks from a birds-eye view. While initially the model struggled to predict consistent paths due to noisy detections received from CenterPoint \cite{yin_center_2021}, adding a small amount of Gaussian noise to the positional observations during training significantly improved the results. The final behavior can be observed in the image as static and dynamic agents can now be reliably predicted. Interestingly, although the majority of generated predictions aligns with the observed actions, the model occasionally exhibits more conservative behavior during road crossings. This phenomenon is highlighted with the yellow and orange trajectories and indicates, that partially displaced observations have a noticeable influence on the models generated velocity profile. Consequently, fine-tuning may be necessary to align with the object detector's standard deviation in estimating object center points. To improve accuracy even further, various data augmentation techniques or an adapted loss-function might be considered.

\subsection{Model performance}
\label{subsec:disc_model_perf}

% features in general
Based on the analysis conducted in \cref{tab:ablation}, the local context provided by the map appears to be the most influential feature. Given Snapshot's accuracy using a simple distance-based selection scheme for both semantic polylines and surrounding agents, the relevance of more sophisticated strategies is questionable. Focusing on semantics alone, considering the closest polylines around the focal agent seems logical and sufficient, as they impact short-term decision-making the most. However, the results in \cref{tab:ablation} suggest that the impact of surrounding agents can be better assessed through collision risk, which is more effective in identifying relevant agents being further away. Although this approach offers a better classification, the distance-based selection provides a good trade-off by being less computationally demanding. Nevertheless, it highlights that networks can derive collision-based cues from positional information alone, suggesting that similar results could be achieved by increasing the number of surrounding agents considered.

\begin{figure}[]
    \centering
    \includegraphics[width=\linewidth]{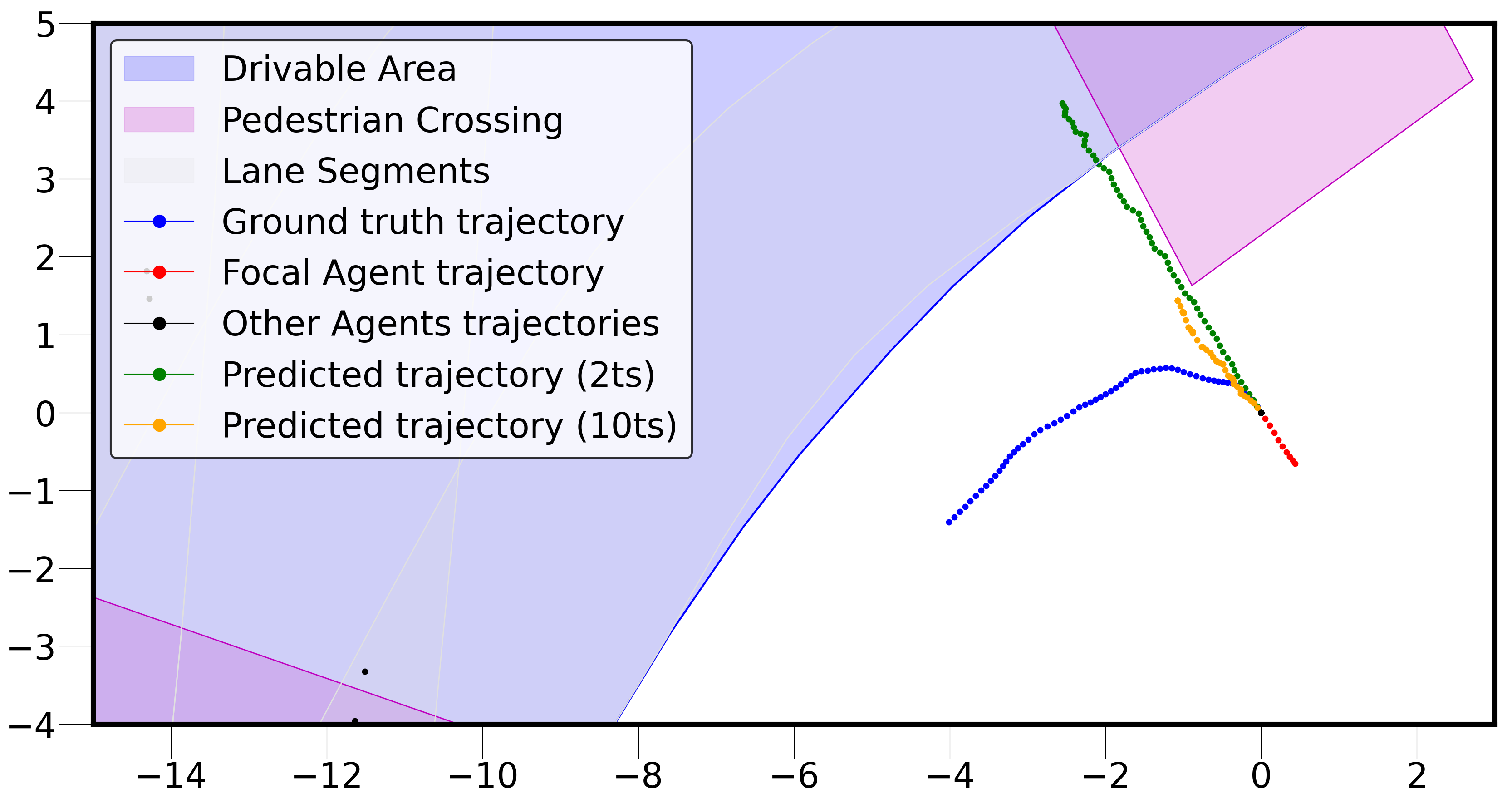}
    \caption{Argoverse 2 scenario visualizing Snapshot's performance for two (green) and ten (yellow) observed timesteps.}
    \label{fig:disc_example_scenarios}
\end{figure}

When analyzing the impact of the motion history as outlined in \cref{fig:res_training}, we find that longer observations positively impact the overall accuracy. These improvements can be attributed to an enhanced scenario understanding as more timesteps provide additional cues to infer a pedestrian's action. This is illustrated in \cref{fig:disc_example_scenarios}, where Snapshot detects a changing movement pattern when using all ten observations. In contrast, with only two observations an anticipated crossing is predicted since only a single velocity can be derived. Despite both predictions not fully capturing the ground truth trajectory, it highlights that an observation length of \SI{1}{\second} is sufficient to differentiate crossing from non-crossing actions which is crucial for practical applications. This finding is supported by \cref{tab:quant_res} where Snapshot outperforms other models employing 50 timesteps on the original validation split. %Therefore, to improve performance further, future research might focus on extending the presented agent selection schemes or explore varying output representations resembling actions rather than trajectories.

\section{Conclusion}
\label{sec:conc}

To address the gap in existing datasets, this work introduces a dedicated pedestrian prediction benchmark based on Argoverse 2, featuring over one million predictable tracks. Building on this foundation, we also developed Snapshot, the first model explicitly designed for urban traffic environments. Utilizing an agent-centric encoding for improved generalization, we employ a simple yet effective input feature representation and a compact architecture to create a scalable model. Notably, this model demonstrates robust performance across varying observation lengths, surpassing the current state of the art while using a five times shorter motion history. As a result, it can predict trajectories in as little as \SI{0.05}{\milli \second} per agent, enabling its real-world applicability that was verified on an automated vehicle.

%%%%%%%%% REFERENCES
{\small
\bibliographystyle{ieee_fullname}
\bibliography{bibs/v0, bibs/datasets, bibs/ext}

\begin{thebibliography}{10}\itemsep=-1pt

\bibitem{achaji_is_2022}
Lina Achaji, Julien Moreau, Thibault Fouqueray, Francois Aioun, and Francois Charpillet.
\newblock Is attention to bounding boxes all you need for pedestrian action prediction?
\newblock In {\em 2022 IEEE Intelligent Vehicles Symposium (IV)}, pages 895--902, 2022.

\bibitem{alahi_social_2016}
Alexandre Alahi, Kratarth Goel, Vignesh Ramanathan, Alexandre Robicquet, Li Fei-Fei, and Silvio Savarese.
\newblock Social {LSTM}: Human trajectory prediction in crowded spaces.
\newblock In {\em 2016 {IEEE} Conference on Computer Vision and Pattern Recognition ({CVPR})}, pages 961--971.
\newblock {ISSN}: 1063-6919.

\bibitem{amirian_social_2019}
Javad Amirian, Jean-Bernard Hayet, and Julien Pettre.
\newblock Social ways: Learning multi-modal distributions of pedestrian trajectories with {GANs}.
\newblock {\em 2019 IEEE/CVF Conference on Computer Vision and Pattern Recognition Workshops (CVPRW)}, pages 2964--2972, 2019.

\bibitem{brscic_person_2013}
Drazen Brscic, Takayuki Kanda, Tetsushi Ikeda, and Takahiro Miyashita.
\newblock Person tracking in large public spaces using 3-d range sensors.
\newblock {\em {IEEE} Transactions on Human-Machine Systems}, 43(6):522--534.

\bibitem{cadena_pedestrian_2022}
Pablo Rodrigo~Gantier Cadena, Yeqiang Qian, Chunxiang Wang, and Ming Yang.
\newblock Pedestrian graph +: A fast pedestrian crossing prediction model based on graph convolutional networks.
\newblock {\em {IEEE} Transactions on Intelligent Transportation Systems}, 23(11):21050--21061.

\bibitem{caesar_nuscenes_2020}
Holger Caesar, Varun Bankiti, Alex~H. Lang, Sourabh Vora, Venice~Erin Liong, Qiang Xu, Anush Krishnan, Yu Pan, Giancarlo Baldan, and Oscar Beijbom.
\newblock {nuScenes}: A multimodal dataset for autonomous driving.
\newblock {\em 2020 IEEE/CVF Conference on Computer Vision and Pattern Recognition (CVPR)}, pages 11618--11628, 2019.

\bibitem{cheng_forecast_2023}
Jie Cheng, Xiaodong Mei, and Ming Liu.
\newblock {Forecast-MAE}: Self-supervised pre-training for motion forecasting with masked autoencoders.
\newblock {\em Proceedings of the IEEE/CVF International Conference on Computer Vision}, 2023.

\bibitem{dosovitskiy_image_2021}
Alexey Dosovitskiy, Lucas Beyer, Alexander Kolesnikov, Dirk Weissenborn, Xiaohua Zhai, Thomas Unterthiner, Mostafa Dehghani, Matthias Minderer, Georg Heigold, Sylvain Gelly, Jakob Uszkoreit, and Neil Houlsby.
\newblock An image is worth 16x16 words: Transformers for image recognition at scale.
\newblock In {\em International Conference on Learning Representations}, 2021.

\bibitem{dutra_gradient-based_2017}
T.~B. Dutra, R. Marques, J.B. Cavalcante-Neto, C.~A. Vidal, and J. Pettré.
\newblock Gradient-based steering for vision-based crowd simulation algorithms.
\newblock {\em Computer Graphics Forum}, 36(2):337--348, 2017.

\bibitem{ettinger_large_2021}
Scott Ettinger, Shuyang Cheng, Benjamin Caine, Chenxi Liu, Hang Zhao, Sabeek Pradhan, Yuning Chai, Ben Sapp, Charles Qi, Yin Zhou, Zoey Yang, Aurelien Chouard, Pei Sun, Jiquan Ngiam, Vijay Vasudevan, Alexander {McCauley}, Jonathon Shlens, and Dragomir Anguelov.
\newblock Large scale interactive motion forecasting for autonomous driving : The waymo open motion dataset.
\newblock In {\em 2021 {IEEE}/{CVF} International Conference on Computer Vision ({ICCV})}, pages 9690--9699. {IEEE}.

\bibitem{fridovich-keil_confidence-aware_2020}
David Fridovich-Keil, Andrea Bajcsy, Jaime~F Fisac, Sylvia~L Herbert, Steven Wang, Anca~D Dragan, and Claire~J Tomlin.
\newblock Confidence-aware motion prediction for real-time collision avoidance1.
\newblock 39(2):250--265.
\newblock Publisher: {SAGE} Publications Ltd {STM}.

\bibitem{gao_vectornet_2020}
Jiyang Gao, Chen Sun, Hang Zhao, Yi Shen, Dragomir Anguelov, Congcong Li, and Cordelia Schmid.
\newblock Vectornet: Encoding hd maps and agent dynamics from vectorized representation.
\newblock In {\em 2020 IEEE/CVF Conference on Computer Vision and Pattern Recognition (CVPR)}, pages 11522--11530, 2020.

\bibitem{gilles_home_2021}
Thomas Gilles, Stefano Sabatini, Dzmitry Tsishkou, Bogdan Stanciulescu, and Fabien Moutarde.
\newblock Home: Heatmap output for future motion estimation.
\newblock In {\em 2021 IEEE International Intelligent Transportation Systems Conference (ITSC)}, pages 500--507, 2021.

\bibitem{golchoubian_pedestrian_2023}
Mahsa Golchoubian, Moojan Ghafurian, Kerstin Dautenhahn, and Nasser~Lashgarian Azad.
\newblock Pedestrian trajectory prediction in pedestrian-vehicle mixed environments: A systematic review.
\newblock {\em {IEEE} Transactions on Intelligent Transportation Systems}, 24(11):11544--11567.

\bibitem{gupta_social_2018}
Agrim Gupta, Justin Johnson, Li Fei-Fei, Silvio Savarese, and Alexandre Alahi.
\newblock Social {GAN}: Socially acceptable trajectories with generative adversarial networks.
\newblock In {\em 2018 IEEE/CVF Conference on Computer Vision and Pattern Recognition (CVPR)}, 2018.

\bibitem{hagenus_survey_2024}
Jeroen Hagenus, Frederik~Baymler Mathiesen, Julian~F. Schumann, and Arkady Zgonnikov.
\newblock A survey on robustness in trajectory prediction for autonomous vehicles.

\bibitem{helbing_social_1995}
Dirk Helbing and Peter Molnar.
\newblock Social force model for pedestrian dynamics.
\newblock {\em Physical review. E, Statistical physics, plasmas, fluids, and related interdisciplinary topics}, 51(5):4282--4286.

\bibitem{kato_autoware_2018}
Shinpei Kato, Shota Tokunaga, Yuya Maruyama, Seiya Maeda, Manato Hirabayashi, Yuki Kitsukawa, Abraham Monrroy, Tomohito Ando, Yusuke Fujii, and Takuya Azumi.
\newblock Autoware on board: Enabling autonomous vehicles with embedded systems.
\newblock In {\em 2018 ACM/IEEE 9th International Conference on Cyber-Physical Systems (ICCPS)}, pages 287--296, 2018.

\bibitem{kim_brvo_2015}
Sujeong Kim, Stephen~J. Guy, Wenxi Liu, David Wilkie, Rynson~W.H. Lau, Ming~C. Lin, and Dinesh Manocha.
\newblock {BRVO}: Predicting pedestrian trajectories using velocity-space reasoning.
\newblock 34(2):201--217.
\newblock Publisher: {SAGE} Publications Ltd {STM}.

\bibitem{koh_modeling_2011}
Wee~Lit Koh and Suiping Zhou.
\newblock Modeling and simulation of pedestrian behaviors in crowded places.
\newblock {\em ACM Trans. Model. Comput. Simul.}, 21(3), 2011.

\bibitem{korbmacher_review_2021}
Raphael Korbmacher and Antoine Tordeux.
\newblock Review of pedestrian trajectory prediction methods: Comparing deep learning and knowledge-based approaches.
\newblock {\em IEEE Transactions on Intelligent Transportation Systems}, 23:24126--24144, 2021.

\bibitem{kosaraju_social-bigat_2019}
Vineet Kosaraju, Amir Sadeghian, Roberto Mart\'{\i}n-Mart\'{\i}n, Ian Reid, S.~Hamid Rezatofighi, and Silvio Savarese.
\newblock Social-bigat: multimodal trajectory forecasting using bicycle-gan and graph attention networks.
\newblock In {\em Proceedings of the 33rd International Conference on Neural Information Processing Systems}, 2019.

\bibitem{kothari_human_2022}
Parth Kothari, Sven Kreiss, and Alexandre Alahi.
\newblock Human trajectory forecasting in crowds: A deep learning perspective.
\newblock 23(7):7386--7400.
\newblock Conference Name: {IEEE} Transactions on Intelligent Transportation Systems.

\bibitem{lan_sept_2023}
Zhiqian Lan, Yuxuan Jiang, Yao Mu, Chen Chen, and Shengbo~Eben Li.
\newblock {SEPT}: Towards efficient scene representation learning for motion prediction.

\bibitem{lerner_crowds_2007}
Alon Lerner, Yiorgos Chrysanthou, and Dani Lischinski.
\newblock Crowds by example.
\newblock {\em Computer Graphics Forum}, 26(3):655--664.

\bibitem{mangalam_goals_2020}
Karttikeya Mangalam, Yang An, Harshayu Girase, and Jitendra Malik.
\newblock From goals, waypoints \& paths to long term human trajectory forecasting.
\newblock {\em 2021 IEEE/CVF International Conference on Computer Vision (ICCV)}, pages 15213--15222, 2020.

\bibitem{miller_magical_1956}
G.~A. Miller.
\newblock The magical number seven plus or minus two: some limits on our capacity for processing information.
\newblock {\em Psychological Review}, 63(2):81--97.

\bibitem{mohamed_social-implicit_2022}
Abduallah Mohamed, Deyao Zhu, Warren Vu, Mohamed Elhoseiny, and Christian Claudel.
\newblock Social-implicit: Rethinking trajectory prediction evaluation and the effectiveness of implicit maximum likelihood estimation.
\newblock In {\em Computer Vision – ECCV 2022: 17th European Conference Proceedings, Part XXII}, page 463–479. Springer-Verlag, 2022.

\bibitem{nayakanti_wayformer_2023}
Nigamaa Nayakanti, Rami Al-Rfou, Aurick Zhou, Kratarth Goel, Khaled~S. Refaat, and Benjamin Sapp.
\newblock Wayformer: Motion forecasting via simple \& efficient attention networks.
\newblock In {\em 2023 {IEEE} International Conference on Robotics and Automation ({ICRA})}, pages 2980--2987.

\bibitem{ondrej_synthetic-vision_2010}
Jan Ondrej, Julien Pettre, Anne-Hélène Olivier, and Stéphane Donikian.
\newblock A synthetic-vision based steering approach for crowd simulation.
\newblock {\em {ACM} Transactions on Graphics}, 29.

\bibitem{pellegrini_youll_2009}
S. Pellegrini, A. Ess, K. Schindler, and L. van Gool.
\newblock You'll never walk alone: Modeling social behavior for multi-target tracking.
\newblock In {\em 2009 {IEEE} 12th International Conference on Computer Vision}, pages 261--268.
\newblock {ISSN}: 2380-7504.

\bibitem{ridel_literature_2018}
Daniela Ridel, Eike Rehder, Martin Lauer, Christoph Stiller, and Denis Wolf.
\newblock A literature review on the prediction of pedestrian behavior in urban scenarios.
\newblock In {\em 2018 21st International Conference on Intelligent Transportation Systems ({ITSC})}, pages 3105--3112. {IEEE}.

\bibitem{robicquet_learning_2016}
Alexandre Robicquet, Amir Sadeghian, Alexandre Alahi, and Silvio Savarese.
\newblock Learning social etiquette: Human trajectory understanding in crowded scenes.
\newblock In Bastian Leibe, Jiri Matas, Nicu Sebe, and Max Welling, editors, {\em Computer Vision – {ECCV} 2016}, Lecture Notes in Computer Science, pages 549--565. Springer International Publishing.

\bibitem{rudenko_thor_2020}
Andrey Rudenko, Tomasz~P. Kucner, Chittaranjan~S. Swaminathan, Ravi~T. Chadalavada, Kai~O. Arras, and Achim~J. Lilienthal.
\newblock {TH{\"O}R}: Human-robot navigation data collection and accurate motion trajectories dataset.
\newblock {\em {IEEE} Robotics and Automation Letters}, 5(2):676--682.

\bibitem{sadeghian_sophie_2018}
Amir Sadeghian, Vineet Kosaraju, Ali Sadeghian, Noriaki Hirose, S.~Hamid Rezatofighi, and Silvio Savarese.
\newblock {SoPhie}: An attentive {GAN} for predicting paths compliant to social and physical constraints.
\newblock {\em 2019 IEEE/CVF Conference on Computer Vision and Pattern Recognition (CVPR)}, pages 1349--1358, 2018.

\bibitem{salzmann_robots_2023}
Tim Salzmann, Lewis Chiang, Markus Ryll, Dorsa Sadigh, Carolina Parada, and Alex Bewley.
\newblock Robots that can see: Leveraging human pose for trajectory prediction.
\newblock {\em {IEEE} Robotics and Automation Letters}, 8(11):7090--7097.

\bibitem{salzmann_trajectron_2021}
Tim Salzmann, Boris Ivanovic, Punarjay Chakravarty, and Marco Pavone.
\newblock Trajectron++: Dynamically-feasible trajectory forecasting with heterogeneous data.
\newblock In {\em European Conference on Computer Vision (ECCV)}, 2020.

\bibitem{scholler_flomo_2021}
Christoph Sch{\"o}ller and Alois Knoll.
\newblock Flomo: Tractable motion prediction with normalizing flows.
\newblock {\em 2021 IEEE/RSJ International Conference on Intelligent Robots and Systems (IROS)}, pages 7977--7984, 2021.

\bibitem{sun_human_nodate}
Jianhua Sun, Yuxuan Li, Liang Chai, Hao-Shu Fang, Yong-Lu Li, and Cewu Lu.
\newblock Human trajectory prediction with momentary observation.
\newblock {\em 2022 IEEE/CVF Conference on Computer Vision and Pattern Recognition (CVPR)}, pages 6457--6466, 2022.

\bibitem{uhlemann_evaluating_2024}
Nico Uhlemann, Felix Fent, and Markus Lienkamp.
\newblock Evaluating pedestrian trajectory prediction methods with respect to autonomous driving.
\newblock {\em IEEE Transactions on Intelligent Transportation Systems}, pages 1--10, 2024.

\bibitem{van_den_berg_reciprocal_2011}
Jur van~den Berg, Stephen~J. Guy, Ming Lin, and Dinesh Manocha.
\newblock Reciprocal n-body collision avoidance.
\newblock In Cédric Pradalier, Roland Siegwart, and Gerhard Hirzinger, editors, {\em Robotics Research}, Springer Tracts in Advanced Robotics, pages 3--19. Springer.

\bibitem{van_den_berg_interactive_2008}
Jur van~den Berg, Sachin Patil, Jason Sewall, Dinesh Manocha, and Ming Lin.
\newblock {\em Interactive navigation of multiple agents in crowded environments}.

\bibitem{vaswani_attention_nodate}
Ashish Vaswani, Noam Shazeer, Niki Parmar, Jakob Uszkoreit, Llion Jones, Aidan~N Gomez, \L~ukasz Kaiser, and Illia Polosukhin.
\newblock Attention is all you need.
\newblock In {\em Advances in Neural Information Processing Systems}, volume~30, 2017.

\bibitem{wang_prophnet_2023}
Xishun Wang, Tong Su, Fang Da, and Xiaodong Yang.
\newblock { ProphNet: Efficient Agent-Centric Motion Forecasting with Anchor-Informed Proposals }.
\newblock In {\em 2023 IEEE/CVF Conference on Computer Vision and Pattern Recognition (CVPR)}, pages 21995--22003, Los Alamitos, CA, USA, 2023. IEEE Computer Society.

\bibitem{wilson_argoverse_2023}
Benjamin Wilson, William Qi, Tanmay Agarwal, John Lambert, Jagjeet Singh, Siddhesh Khandelwal, Bowen Pan, Ratnesh Kumar, Andrew Hartnett, Jhony~Kaesemodel Pontes, Deva Ramanan, Peter Carr, and James Hays.
\newblock Argoverse 2: Next generation datasets for self-driving perception and forecasting.
\newblock In {\em Proceedings of the Neural Information Processing Systems Track on Datasets and Benchmarks (NeurIPS Datasets and Benchmarks 2021)}, 2021.

\bibitem{yin_center_2021}
Tianwei Yin, Xingyi Zhou, and Philipp Kr{\"a}henb{\"u}hl.
\newblock Center-based 3d object detection and tracking.
\newblock {\em 2021 IEEE/CVF Conference on Computer Vision and Pattern Recognition (CVPR)}, pages 11779--11788, 2020.

\bibitem{yuan_agentformer_2021}
Ye Yuan, Xinshuo Weng, Yanglan Ou, and Kris Kitani.
\newblock {AgentFormer}: Agent-aware transformers for socio-temporal multi-agent forecasting.
\newblock {\em 2021 IEEE/CVF International Conference on Computer Vision (ICCV)}, pages 9793--9803, 2021.

\bibitem{yue_human_2022}
Jiangbei Yue, Dinesh Manocha, and He Wang.
\newblock Human trajectory prediction via neural social physics.
\newblock In {\em European Conference on Computer Vision (ECCV)}, 2022.

\bibitem{zamboni_pedestrian_2022}
Simone Zamboni, Zekarias~Tilahun Kefato, Sarunas Girdzijauskas, Christoffer Norén, and Laura Dal~Col.
\newblock Pedestrian trajectory prediction with convolutional neural networks.
\newblock {\em Pattern Recognition}, 121:108252.

\bibitem{zhang_simpl_2024}
Lu Zhang, Peiliang Li, Sikang Liu, and Shaojie Shen.
\newblock Simpl: A simple and efficient multi-agent motion prediction baseline for autonomous driving.
\newblock {\em IEEE Robotics and Automation Letters}, 9(4):3767--3774, 2024.

\bibitem{zhang_forceformer_2023}
Weicheng Zhang, Hao Cheng, Fatema~T. Johora, and Monika Sester.
\newblock {ForceFormer}: Exploring social force and transformer for pedestrian trajectory prediction.
\newblock In {\em 2023 {IEEE} 35th Symposium on Intelligent Vehicles (IV)}.

\bibitem{zhang_real-time_2023}
Zhejun Zhang, Alexander Liniger, Christos Sakaridis, Fisher Yu, and Luc Van~Gool.
\newblock Real-time motion prediction via heterogeneous polyline transformer with relative pose encoding.
\newblock In {\em Advances in Neural Information Processing Systems (NeurIPS)}, 2023.

\bibitem{zhou_query-centric_2023}
Zikang Zhou, Jianping Wang, Yung–Hui Li, and Yu–Kai Huang.
\newblock Query-centric trajectory prediction.
\newblock In {\em 2023 {IEEE}/{CVF} Conference on Computer Vision and Pattern Recognition ({CVPR})}, pages 17863--17873. {IEEE}.

\bibitem{zhou_hivt_2022}
Zikang Zhou, Luyao Ye, Jianping Wang, Kui Wu, and Kejie Lu.
\newblock {HiVT}: Hierarchical vector transformer for multi-agent motion prediction.
\newblock In {\em 2022 {IEEE}/{CVF} Conference on Computer Vision and Pattern Recognition ({CVPR})}, pages 8813--8823. {IEEE}.

\end{thebibliography}
}

\end{document}